\documentclass[11pt,a4paper]{article}
\usepackage[hyperref]{emnlp-ijcnlp-2019}
\usepackage{times}
\usepackage{latexsym}
\usepackage{cleveref}
\usepackage{graphicx}
\usepackage{changepage}
\usepackage{enumitem}
\usepackage{booktabs}
\usepackage{multicol}
\usepackage{float}

\newlength{\offsetpage}
\newenvironment{widepage}{\begin{adjustwidth}{-\offsetpage}{-\offsetpage}%
    \addtolength{\textwidth}{2\offsetpage}}%
{\end{adjustwidth}}

\usepackage{url}

\aclfinalcopy 

\title{How Much Knowledge Can You Pack \\ Into the Parameters of a Language Model?}

\author{Adam Roberts\thanks{$\;\;$Equal contribution. Noam suggested trying T5 on open-domain QA and coded and ran initial experiments on TriviaQA showing improved performance with model size. Adam wrote the code and ran most experiments.
Colin set the research scope, wrote the paper, and ran a few experiments.
} \\
  Google \\
  {\tt adarob@google.com} \\\And
  Colin Raffel$^*$ \\
  Google \\
  {\tt craffel@gmail.com} \\\And
  Noam Shazeer \\
  Google \\
  {\tt noam@google.com} \\}

\date{}

\begin{document}
\maketitle
\begin{abstract}
It has recently been observed that neural language models trained on unstructured text can implicitly store and retrieve knowledge using natural language queries.
In this short paper, we measure the practical utility of this approach by fine-tuning pre-trained models to answer questions \textit{without access to any external context or knowledge}.
We show that this approach scales with model size and performs competitively with open-domain systems that explicitly retrieve answers from an external knowledge source when answering questions.
To facilitate reproducibility and future work, we release
our code and trained models.\footnote{\url{https://goo.gle/t5-cbqa}}
\end{abstract}

\section{Introduction}

Big, deep neural language models that have been pre-trained on unlabeled text have proven to be extremely performant when fine-tuned on downstream Natural Language Processing (NLP) tasks \cite{devlin2018bert,yang2019xlnet,liu2019roberta,lan2019albert,raffel2019exploring}.
Interestingly, it has also recently been observed that these models can internalize a sort of implicit ``knowledge base'' after pre-training \cite{petroni2019language,jiang2019can,talmor2019olmpics}.
This behavior is potentially useful because
1) the knowledge is built up by pre-training on unstructured and unlabeled text data, which is freely available in huge quantities on the Internet \cite{raffel2019exploring,wenzek2019ccnet}, and
2) it is possible to retrieve information using informal natural language queries since these pre-trained language models excel when fine-tuned on natural language understanding tasks.

\begin{figure}
    \centering
    \begin{widepage}
    \includegraphics[width=1.2\columnwidth]{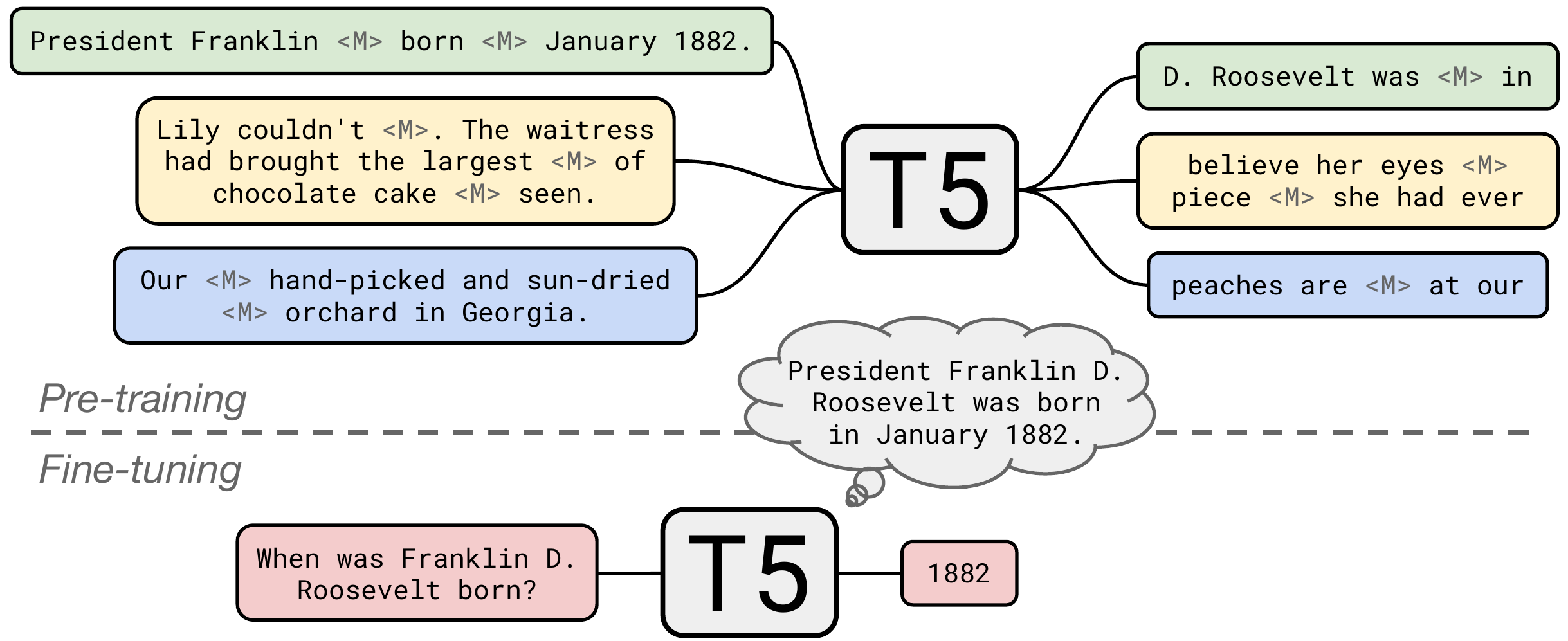}
    \end{widepage}
    \caption{T5 is pre-trained to fill in dropped-out spans of text (denoted by \texttt{<M>}) from documents in a large, unstructured text corpus. We fine-tune T5 to answer questions without inputting any additional information or context. This forces T5 to answer questions based on ``knowledge'' that it internalized during pre-training.}
    \label{fig:my_label}
\end{figure}

Past work investigating ``language models as knowledge bases'' has typically tried to understand the scope of the information stored in the model using synthetic tasks that are similar to the pre-training objective \cite{petroni2019language,jiang2019can} and/or measure reasoning capabilities \cite{talmor2019olmpics}.
In this work, we take a different approach by evaluating the capability of language models on the practical task of open-domain question answering -- specifically, we fine-tune the model to answer questions \textit{without access to any external knowledge or context}.
To do so, the model must parse a natural language query and ``look up information'' stored in its parameters.

Most past work on question answering either explicitly feeds pertinent information to the model alongside the question (for example, an article that contains the answer \cite{rajpurkar2016squad,zhang2018record,khashabi2018looking,clark2019boolq}) or allows the model to retrieve information from an external knowledge source \cite{berant2013semantic,chen2017reading}.
By feeding the model the input question alone, we can determine how much knowledge it has stored in its parameters while measuring its performance on a useful real-world problem.
We refer to this task as ``closed-book question answering''.

A separate question we address in this work is whether models with more parameters end up storing more information.
It has been shown that transfer learning performance on many downstream tasks tends to improve as the model size and amount of unsupervised pre-training increases \cite{radford2019language,liu2019roberta,raffel2019exploring}.
In this work, we leverage the pre-trained ``T5'' models released by \citet{raffel2019exploring}, the largest of which has around 11 billion parameters.
By measuring knowledge retrieval capabilities on models of various sizes, including models that have an order of magnitude more parameters than considered in past work, we can explore how well our approach scales.


\section{Background}

\paragraph{Question Answering}
The task of training a model to either select or output the correct answer to a given question is referred to as ``question answering''.
The most popular variant of this task feeds the model some ``context'' containing the answer (for example, a paragraph from an encyclopedia article) alongside the question \cite{rajpurkar2016squad,zhang2018record,khashabi2018looking,clark2019boolq}.
Models can be trained either to indicate the span of the context that contains the answer or output the text of the answer itself.
Since this format can be seen as reading some text and answering a question about it, it has been referred to as ``reading comprehension''.

A more difficult variant is ``open-domain question answering'' \cite{prager2006open}, where the model can be asked arbitrary context-independent questions (e.g.\ well-known facts or historical details).
It is typically assumed that the model can access an external collection of knowledge when answering questions (e.g.\ a structured knowledge base or unstructured text corpus), but the model is not given any information about where in the collection the answer appears.
The reading comprehension task can be considered a simplified version of open-domain question answering where the model is provided with the oracle context to answer a given question.
As an analogy, the open-domain question answering system acts as if it is taking an \textbf{open-book} exam where it can find and use information in an external source of knowledge.\footnote{While our definition of open-book is the same as in the OpenBookQA dataset introduced by \citet{OpenBookQA2018}, we do not directly address multi-hop inference in this work.}

In this work, we consider open-domain question answering with the additional constraint that the model is \textit{not} allowed to access any external knowledge whatsoever when answering questions.
Instead, the model itself must be pre-trained to store knowledge in its parameters before being fine-tuned to answer questions.
In one view, this can be seen as an alternative way to approach open-domain question answering where instead of learning to access external knowledge the model needs to have ``memorized'' it in order to answer questions; in another view, this constraint creates a third and potentially more ambitious variant of the question answering task.
A model that answers questions in this way is metaphorically similar to a student taking a \textbf{closed-book} exam, where the student must study and memorize all pertinent information before taking the test.

\paragraph{Transfer Learning with Language Models}
In the past few years, it has become increasingly common to pre-train a language model using an unsupervised objective on a large, unstructured text corpus before fine-tuning it on a downstream task of interest \cite{dai2015semi,howard2018universal,radford2018improving}.
The popularity of this form of ``transfer learning'' is attributable to its empirical success on many NLP tasks \cite{peters2018deep,devlin2018bert,yang2019xlnet,lan2019albert,raffel2019exploring}.
Loosely speaking, the pre-training step may provide the model with some generally-useful awareness of meaning, syntax, and ``world knowledge''.
In question answering in particular, most state-of-the-art systems use some form of transfer learning.

Currently, the most popular model architectures used in transfer learning for NLP are Transformer-based \cite{vaswani2017attention} ``encoder-only'' models like BERT \cite{devlin2018bert}.
These models can produce a single prediction for each input token and have been applied to reading comprehension-style question answering by predicting which tokens of the context contain the answer.
Encoder-only models are not applicable to closed-book question answering because no context is provided to extract the answer span from.
An alternative to encoder-only models, recently advocated by \citet{raffel2019exploring}, is to treat every NLP task as a text-to-text problem using an encoder-decoder Transformer.
When this framework is applied to question answering, the model is trained to generate the literal text of the answer in a free-form fashion.
Despite the potential difficulty of generating rather than extracting the answer, \citet{raffel2019exploring} demonstrated state-of-the-art results on the SQuAD \cite{rajpurkar2016squad}, MultiRC \cite{khashabi2018looking}, BoolQ \cite{clark2019boolq}, and ReCoRD \cite{zhang2018record} reading comprehension tasks.

The text-to-text framework is directly applicable to closed-book question answering since the model can be trained to generate an answer with or without any additional information in its input.
Crucially, fine-tuning a text-to-text model to answer questions without any context requires that the model retrieve information from its parameters that it learned during pre-training.
\citet{radford2019language} considered a similar task to evaluate the zero-shot question answering capabilities of a language model.
The concurrent ``RELIC'' and ``EAE'' models of \citet{ling2020learning} and \citet{fevry2020eae} learn representations for an explicitly predefined set of entities and are evaluated on the same closed-book variant of TriviaQA that we consider.
Relatedly, \citet{petroni2019language} show that it is possible to manually convert some questions to a fill-in-the-blank format amenable to an encoder-only model (e.g.\ ``Who developed the theory of relativity?''\ gets mapped to ``The theory of relativity was developed by \_\_\_\_'').

\section{Experiments}
\label{sec:experiments}

\paragraph{Datasets}
We consider the following open-domain question answering datasets:
\textit{Natural Questions} \cite{kwiatkowski2019natural}, a dataset of questions from web queries, each accompanied by a Wikipedia article containing the answer;
\textit{WebQuestions} \cite{berant2013semantic}, comprising questions from web queries matched to corresponding entries in FreeBase \cite{bollacker2008freebase};
and \textit{TriviaQA} \cite{joshi2017triviaqa}, a collection of questions from quiz league websites where each question is accompanied by pages from web and Wikipedia searches that may contain the answer.
In this work, we only make use of the questions from each dataset -- \textit{we completely ignore the matching documents supplied for each question}.

For WebQuestions and TriviaQA we follow the standard evaluation procedures where each predicted answer is compared to the ground-truth after both are lowercased and stripped of articles, punctuation, and duplicate whitespace \cite{rajpurkar2016squad}.
For Natural Questions, we evaluate using both 1) the standard ``open-domain'' version as used e.g.\ by \cite{lee2019latent,min2019knowledge,min2019discrete,asai2019learning} where the model is only required to produce a single normalized answer and 2) the standard multi-answer variant used with reading comprehension systems \cite{kwiatkowski2019natural}.
We review the details of Natural Questions evaluation in \cref{sec:nqeval}.

Note that Natural Questions and TriviaQA have private test sets, so standard practice on their open-domain variants is to report performance on the development sets.
However, we also include our results on the official TriviaQA test set by fine-tuning on the unfiltered training set and submitting our test set predictions to the leaderboard for the Wikipedia domain.
We urge future work to adopt this approach to help ensure the validity of results and avoid potentially overfitting to a public set.

\paragraph{Training}

We leverage the pre-trained models provided by \citet{raffel2019exploring}, referred to as the ``Text-to-Text Transfer Transformer'' (T5).
The original T5 models were pre-trained on a multitask mixture including an unsupervised ``span corruption'' task on the C4 dataset as well as supervised translation, summarization, classification, and reading comprehension tasks.
Note that none of the reading comprehension datasets used for pre-training T5 overlap with the question answering datasets that we consider in this paper.
In order to measure how performance scales with model size, we perform experiments with the Base (220 million parameters), Large (770 million), 3B (3 billion), and 11B (11 billion) variants of T5.
Given that the T5 models were pre-trained on a multitask mixture including question answering, we also report performance using the ``T5.1.1'' checkpoints, which were pre-trained on unlabeled data only.\footnote{\url{https://goo.gle/t5-checkpoints}}

For fine-tuning the T5 checkpoints, we follow the procedure used in \citet{raffel2019exploring} without any additional hyperparameter tuning:
We use the AdaFactor optimizer \cite{shazeer2018adafactor} with a constant learning rate of $0.001$, $10\%$ dropout rate, and a batch size of $196{,}608$ tokens.
We halve the batch and double the dropout rate for WebQuestions due to its small size.
For the T5.1.1 checkpoints, we follow the same procedure but with a dropout rate of $5\%$ for all three datasets.

For evaluation, we follow the procedure used in \citet{lee2019latent}: for each dataset, we hold out $10\%$ of the training set as a validation split, fine-tune a model from the remaining 90\% of examples, and select the best-performing checkpoint for final evaluation on the test set. While we chose to train for $20{,}000$ steps,
our validation accuracy typically plateaued after only a few hundred steps and showed no signs of overfitting.

We decode the model's predictions by choosing the most likely token at each timestep.
To map question answering tasks to the text-to-text format, we simply feed the question with a task-specific prefix into the model as input and train it to predict the literal answer text as output.

\paragraph{Salient Span Masking}

Recently, \citet{guu2020realm} found that a ``salient span masking'' (SSM) pre-training objective produced substantially better results in open-domain question answering.
This approach first uses BERT \cite{devlin2018bert} to mine sentences that contain salient spans (named entities and dates) from Wikipedia.
The question answering model is then pre-trained to reconstruct masked-out spans from these sentences, which \citet{guu2020realm} hypothesize helps the model ``focus on problems that require world knowledge''.
We experimented with using the same SSM data and objective to continue pre-training the T5 checkpoints for $100{,}000$ additional steps before fine-tuning for question answering.

\paragraph{Results}

\begin{table}[h!]
    \centering
    \footnotesize
    \caption{Scores achieved by fine-tuning T5 on the open-domain Natural Questions (NQ), WebQuestions (WQ), and TriviaQA (TQA) tasks.
    }
    \label{tab:results}
    \begin{tabular}{lcccc}
    \toprule
        & NQ & WQ & \multicolumn{2}{c}{TQA} \\
        &    &    & dev  & test \\
        \cmidrule(l{3pt}r{3pt}){1-1} \cmidrule(l{3pt}r{3pt}){2-2}  \cmidrule(l{3pt}r{3pt}){3-3} \cmidrule(l{3pt}r{3pt}){4-4} \cmidrule(l{3pt}r{3pt}){5-5}
        \citet{chen2017reading}  & --   & 20.7 & --   & --  \\
        \citet{lee2019latent}    & 33.3 & 36.4 & 47.1 & -- \\
        \citet{min2019discrete}  & 28.1 & --   & 50.9 & -- \\
        \citet{min2019knowledge} & 31.8 & 31.6 & 55.4 & -- \\
        \citet{asai2019learning} & 32.6 & --   & --   & --  \\
        \citet{ling2020learning} & --   & --   & 35.7 & -- \\
        \citet{guu2020realm}     & 40.4 & 40.7 & --   & --   \\
        \citet{fevry2020eae}     & --   & --   & 43.2 & 53.4 \\
        \citet{karpukhin2020dpr} & \textbf{41.5}  & 42.4  & \textbf{57.9} & -- \\
        \cmidrule(l{3pt}r{3pt}){1-1} \cmidrule(l{3pt}r{3pt}){2-2}  \cmidrule(l{3pt}r{3pt}){3-3} \cmidrule(l{3pt}r{3pt}){4-4} \cmidrule(l{3pt}r{3pt}){5-5}
        T5-Base                  & 25.9 & 27.9 & 23.8 & 29.1  \\
        T5-Large                 & 28.5 & 30.6& 28.7 & 35.9  \\
        T5-3B                    & 30.4 & 33.6 & 35.1 & 43.4  \\
        T5-11B                   & 32.6 & 37.2 & 42.3 & 50.1  \\
        \cmidrule(l{3pt}r{3pt}){1-1} \cmidrule(l{3pt}r{3pt}){2-2}  \cmidrule(l{3pt}r{3pt}){3-3} \cmidrule(l{3pt}r{3pt}){4-4} \cmidrule(l{3pt}r{3pt}){5-5}
        T5-11B + SSM             & 34.8 & 40.8 & 51.0 & 60.5 \\
        \cmidrule(l{3pt}r{3pt}){1-1} \cmidrule(l{3pt}r{3pt}){2-2}  \cmidrule(l{3pt}r{3pt}){3-3} \cmidrule(l{3pt}r{3pt}){4-4} \cmidrule(l{3pt}r{3pt}){5-5}
        T5.1.1-Base                 & 25.7 & 28.2 & 24.2 & 30.6 \\
        T5.1.1-Large                & 27.3 & 29.5 & 28.5 & 37.2 \\
        T5.1.1-XL                   & 29.5 & 32.4 & 36.0 & 45.1 \\
        T5.1.1-XXL                  & 32.8 & 35.6 & 42.9 & 52.5 \\
        \cmidrule(l{3pt}r{3pt}){1-1} \cmidrule(l{3pt}r{3pt}){2-2}  \cmidrule(l{3pt}r{3pt}){3-3} \cmidrule(l{3pt}r{3pt}){4-4} \cmidrule(l{3pt}r{3pt}){5-5}
        T5.1.1-XXL + SSM             & 35.2 & \textbf{42.8} & 51.9 & \textbf{61.6} \\
        \bottomrule
    \end{tabular}
\end{table}

Our results on the open-domain Natural Questions, WebQuestions, and TriviaQA tasks are shown in \cref{tab:results}.
Notably, performance on each dataset improves as the model size increases, with either T5-11B or the comparably-sized T5.1.1-XXL (pre-trained only on unlabeled data) performing best in every case.
Further, we find that using \citet{guu2020realm}'s SSM pre-training produces a substantial boost in performance.
T5.1.1-XXL with SSM ultimately achieves state-of-the-art on WebQuestions and our largest models beat most other methods on Natural Questions and TriviaQA.
Importantly, all previous methods except \citet{ling2020learning} and \citet{fevry2020eae} operate in the ``open-book'' setting by explicitly retrieving and using information from an external knowledge source.
While our largest models are computationally intensive, we note that most open-domain question answering systems must first do an expensive lookup step over the entire knowledge corpus and then attend to a long document to extract an answer.
Our approach omits both of these steps, which ultimately saves a large amount of computation and memory.


Having established that our approach is competitive on open-domain question answering, we now evaluate it on the standard (and more difficult) multi-answer variant of Natural Questions.
Virtually all models used on this task are reading comprehension systems that select the correct answer from an oracle context.
After fine-tuning, T5-11B + SSM achieves a recall of $36.2$ on the validation set, which lags behind the state-of-the-art score of $51.9$ from \citet{pan2019frustratingly}\footnote{Validation set recall scores from \citet{pan2019frustratingly} were reported in private correspondence with the authors.} but outperforms the best baseline published alongside the dataset (recall of $33.2$ \cite{kwiatkowski2019natural}).
This shows that T5 can effectively answer questions with multiple answers.
We discuss additional experiments and negative results in \cref{sec:negative}.

\begin{table*}[h!t]
    \centering
    \footnotesize
    \caption{A breakdown of the 150 hand-evaluated examples from Natural Questions where the T5 predictions were labelled as incorrect by the automatic procedure. We found only $62\%$ of these to be true positives.}
    \label{tab:nqeval-manual}
    \begin{tabular}{lrp{4.2cm}lp{2.3cm}}
    \toprule
        && \multicolumn{3}{c}{Example}\\
        \cmidrule(lr){3-5} 
        Category & Percentage & Question & Target(s) & T5 Prediction \\
        \midrule
        True Negative & $62.0\%$ & what does the ghost of christmas present sprinkle from his torch & little warmth, warmth & confetti \\
        Phrasing Mismatch  & $13.3\%$ & who plays red on orange is new black & kate mulgrew & katherine kiernan maria mulgrew\\
        Incomplete Annotation & $13.3\%$ &  where does the us launch space shuttles from & florida & kennedy lc39b \\
        Unanswerable  & $11.3\%$ & who is the secretary of state for northern ireland & karen bradley & james brokenshire \\
        \bottomrule
    \end{tabular}
\end{table*}

\paragraph{Human Evaluation}


The benchmarks we used and the ``exact match'' score assume that the model directly extracts answers from an external knowledge source.
In contrast, our model generates answers in a free-form fashion.
We hypothesize that this results in many false negatives when answers do not exactly match the ground-truth context intended for each question.
We therefore manually inspected $150$ examples from the Natural Questions validation set where our model's prediction was counted as incorrect in hopes of identifying ``false negatives'' according to the exact match metric.
We found that false negatives fell into three broad categories:
First, answers with meaning-preserving differences in phrasing (e.g.\ ``April 15'' vs. ``April 15th'');
second, questions that were missing all possible correct answers (e.g.\ ``where does the us launch space shuttles from'' was annotated with the single ground-truth answer ``florida'', despite many possible correct answers such as ``Kennedy Space Center'', ``Merritt Island'', ``Cape Canaveral'', etc.);
and finally, some questions were unanswerable without knowing the exact time or article they referred to (e.g.\ ``what is the latest version of microsoft office 2010'' depends on when the question is being asked).
We provide examples of each of these false negative types in \cref{tab:nqeval-manual}.
We note that open-book question answering systems could also be impacted to a lesser extent by these issues (e.g.\ if they select a slightly different answer span from the annotated one or retrieve a non-golden document that contains a different correct answer).

Of the $150$ examples inspected, we found that $20$ were marked as incorrect due to differences in phrasing, another $20$ were not annotated with all correct answers, and $17$ were unanswerable without appropriate context.
Removing unanswerable questions from the validation set and recomputing our model's accuracy based on this false-negative rate produces a score of $57.8$.
This suggests that the performance of closed-book question answering systems (in terms of how often it correctly answers questions) is substantially underestimated by the evaluation procedure used in these benchmarks.
For full transparency, we publicly release the results of our human evaluation  and include an appropriate reference when we determined that a predicted answer was missing from ground-truth.\footnote{\url{https://goo.gle/t5-cbqa-human-eval}}



\section{Conclusion}
In this short paper, we have shown that large language models pre-trained on unstructured text can attain competitive results on open-domain question answering benchmarks without any access to external knowledge.
This suggests a fundamentally different approach to designing question answering systems, motivating many threads for future work:
First, we obtained state-of-the-art results only with the largest models which had around 11 billion parameters.
This model size can be prohibitively expensive in resource-constrained settings, prompting future work on more efficient language models.
Second, ``open-book'' models typically provide some indication of what information they accessed when answering a question.
This can provide a useful form of interpretability.
In contrast, our model distributes knowledge in its parameters in an inexplicable way and hallucinates realistic-looking answers when it is unsure.
Third, the maximum-likelihood objective used to train our model provides no guarantees as to whether a model will learn a fact or not.
This makes it difficult to ensure that the model obtains specific knowledge over the course of pre-training and prevents us from explicitly updating or removing knowledge from a pre-trained model.
Finally, the tasks we used in this paper mainly measure ``trivia''-style knowledge.
We are therefore interested in measuring performance on question answering tasks that require reasoning capabilities such as DROP \cite{dua2019drop}.

\newpage

\section*{Acknowledgments}

We thank Kelvin Guu, Kenton Lee, Ming-Wei Chang, Zora Tung, and Ice Pasupat for providing the open-domain question answering evaluation setup and access to their salient span-annotated data; Roy Frostig and Katherine Lee for comments and suggestions on this manuscript; Noah Constant for suggesting we try salience span masking; and Monica Dinculescu for building an interactive demonstration of our results.\footnote{\url{http://t5-trivia.glitch.me/}}

\bibliography{emnlp-ijcnlp-2019}

\begin{thebibliography}{37}
\expandafter\ifx\csname natexlab\endcsname\relax\def\natexlab#1{#1}\fi

\bibitem[{Asai et~al.(2019)Asai, Hashimoto, Hajishirzi, Socher, and
  Xiong}]{asai2019learning}
Akari Asai, Kazuma Hashimoto, Hannaneh Hajishirzi, Richard Socher, and Caiming
  Xiong. 2019.
\newblock Learning to retrieve reasoning paths over {Wikipedia} graph for
  question answering.
\newblock \emph{arXiv preprint arXiv:1911.10470}.

\bibitem[{Berant et~al.(2013)Berant, Chou, Frostig, and
  Liang}]{berant2013semantic}
Jonathan Berant, Andrew Chou, Roy Frostig, and Percy Liang. 2013.
\newblock Semantic parsing on freebase from question-answer pairs.
\newblock In \emph{Proceedings of the 2013 Conference on Empirical Methods in
  Natural Language Processing}.

\bibitem[{Bollacker et~al.(2008)Bollacker, Evans, Paritosh, Sturge, and
  Taylor}]{bollacker2008freebase}
Kurt Bollacker, Colin Evans, Praveen Paritosh, Tim Sturge, and Jamie Taylor.
  2008.
\newblock Freebase: a collaboratively created graph database for structuring
  human knowledge.
\newblock In \emph{Proceedings of the 2008 ACM SIGMOD International Conference
  on Management of Data}, pages 1247--1250.

\bibitem[{Chen et~al.(2017)Chen, Fisch, Weston, and Bordes}]{chen2017reading}
Danqi Chen, Adam Fisch, Jason Weston, and Antoine Bordes. 2017.
\newblock Reading {Wikipedia} to answer open-domain questions.
\newblock \emph{arXiv preprint arXiv:1704.00051}.

\bibitem[{Clark et~al.(2019)Clark, Lee, Chang, Kwiatkowski, Collins, and
  Toutanova}]{clark2019boolq}
Christopher Clark, Kenton Lee, Ming-Wei Chang, Tom Kwiatkowski, Michael
  Collins, and Kristina Toutanova. 2019.
\newblock {BoolQ}: Exploring the surprising difficulty of natural yes/no
  questions.
\newblock \emph{arXiv preprint arXiv:1905.10044}.

\bibitem[{Dai and Le(2015)}]{dai2015semi}
Andrew~M. Dai and Quoc~V. Le. 2015.
\newblock Semi-supervised sequence learning.
\newblock In \emph{Advances in Neural Information Processing Systems}.

\bibitem[{Devlin et~al.(2018)Devlin, Chang, Lee, and
  Toutanova}]{devlin2018bert}
Jacob Devlin, Ming-Wei Chang, Kenton Lee, and Kristina Toutanova. 2018.
\newblock {BERT}: Pre-training of deep bidirectional transformers for language
  understanding.
\newblock \emph{arXiv preprint arXiv:1810.04805}.

\bibitem[{Dua et~al.(2019)Dua, Wang, Dasigi, Stanovsky, Singh, and
  Gardner}]{dua2019drop}
Dheeru Dua, Yizhong Wang, Pradeep Dasigi, Gabriel Stanovsky, Sameer Singh, and
  Matt Gardner. 2019.
\newblock Drop: A reading comprehension benchmark requiring discrete reasoning
  over paragraphs.
\newblock \emph{arXiv preprint arXiv:1903.00161}.

\bibitem[{F{\'e}vry et~al.(2020)F{\'e}vry, Soares, FitzGerald, Choi, and
  Kwiatkowski}]{fevry2020eae}
Thibault F{\'e}vry, Livio~Baldini Soares, Nicholas FitzGerald, Eunsol Choi, and
  Tom Kwiatkowski. 2020.
\newblock Entities as experts: Sparse memory access with entity supervision.
\newblock \emph{arXiv preprint arXiv:2004.07202}.

\bibitem[{Guu et~al.(2020)Guu, Lee, Tung, Panupong, and Chang}]{guu2020realm}
Kelvin Guu, Kenton Lee, Zora Tung, Pasupat Panupong, and Ming-Wei Chang. 2020.
\newblock Realm: Retrieval-augmented language model pre-training.
\newblock \emph{arXiv preprint arXiv:2002.08909}.

\bibitem[{Howard and Ruder(2018)}]{howard2018universal}
Jeremy Howard and Sebastian Ruder. 2018.
\newblock Universal language model fine-tuning for text classification.
\newblock \emph{arXiv preprint arXiv:1801.06146}.

\bibitem[{Jiang et~al.(2019)Jiang, Xu, Araki, and Neubig}]{jiang2019can}
Zhengbao Jiang, Frank~F Xu, Jun Araki, and Graham Neubig. 2019.
\newblock How can we know what language models know?
\newblock \emph{arXiv preprint arXiv:1911.12543}.

\bibitem[{Joshi et~al.(2017)Joshi, Choi, Weld, and
  Zettlemoyer}]{joshi2017triviaqa}
Mandar Joshi, Eunsol Choi, Daniel~S. Weld, and Luke Zettlemoyer. 2017.
\newblock {TriviaQA}: A large scale distantly supervised challenge dataset for
  reading comprehension.
\newblock \emph{arXiv preprint arXiv:1705.03551}.

\bibitem[{Karpukhin et~al.(2020)Karpukhin, Ouguz, Min, Wu, Edunov, Chen, and
  tau Yih}]{karpukhin2020dpr}
Vladimir Karpukhin, Barlas Ouguz, Sewon Min, Ledell~Yu Wu, Sergey Edunov, Danqi
  Chen, and Wen tau Yih. 2020.
\newblock Dense passage retrieval for open-domain question answering.
\newblock \emph{arXiv preprint arXiv:2004.04906}.

\bibitem[{Khashabi et~al.(2018)Khashabi, Chaturvedi, Roth, Upadhyay, and
  Roth}]{khashabi2018looking}
Daniel Khashabi, Snigdha Chaturvedi, Michael Roth, Shyam Upadhyay, and Dan
  Roth. 2018.
\newblock Looking beyond the surface: A challenge set for reading comprehension
  over multiple sentences.
\newblock In \emph{Proceedings of North American Chapter of the Association for
  Computational Linguistics (NAACL)}.

\bibitem[{Kwiatkowski et~al.(2019)Kwiatkowski, Palomaki, Redfield, Collins,
  Parikh, Alberti, Epstein, Polosukhin, Devlin, Lee
  et~al.}]{kwiatkowski2019natural}
Tom Kwiatkowski, Jennimaria Palomaki, Olivia Redfield, Michael Collins, Ankur
  Parikh, Chris Alberti, Danielle Epstein, Illia Polosukhin, Jacob Devlin,
  Kenton Lee, et~al. 2019.
\newblock Natural questions: a benchmark for question answering research.
\newblock \emph{Transactions of the Association for Computational Linguistics},
  7.

\bibitem[{Lan et~al.(2019)Lan, Chen, Goodman, Gimpel, Sharma, and
  Soricut}]{lan2019albert}
Zhenzhong Lan, Mingda Chen, Sebastian Goodman, Kevin Gimpel, Piyush Sharma, and
  Radu Soricut. 2019.
\newblock {ALBERT}: A lite {BERT} for self-supervised learning of language
  representations.
\newblock \emph{arXiv preprint arXiv:1909.11942}.

\bibitem[{Lee et~al.(2019)Lee, Chang, and Toutanova}]{lee2019latent}
Kenton Lee, Ming-Wei Chang, and Kristina Toutanova. 2019.
\newblock Latent retrieval for weakly supervised open domain question
  answering.
\newblock \emph{arXiv preprint arXiv:1906.00300}.

\bibitem[{Ling et~al.(2020)Ling, FitzGerald, Shan, Soares, F{\'e}vry, Weiss,
  and Kwiatkowski}]{ling2020learning}
Jeffrey Ling, Nicholas FitzGerald, Zifei Shan, Livio~Baldini Soares, Thibault
  F{\'e}vry, David Weiss, and Tom Kwiatkowski. 2020.
\newblock Learning cross-context entity representations from text.
\newblock \emph{arXiv preprint arXiv:2001.03765}.

\bibitem[{Liu et~al.(2019)Liu, Ott, Goyal, Du, Joshi, Chen, Levy, Lewis,
  Zettlemoyer, and Stoyanov}]{liu2019roberta}
Yinhan Liu, Myle Ott, Naman Goyal, Jingfei Du, Mandar Joshi, Danqi Chen, Omer
  Levy, Mike Lewis, Luke Zettlemoyer, and Veselin Stoyanov. 2019.
\newblock {RoBERTa}: A robustly optimized {BERT} pretraining approach.
\newblock \emph{arXiv preprint arXiv:1907.11692}.

\bibitem[{Mihaylov et~al.(2018)Mihaylov, Clark, Khot, and
  Sabharwal}]{OpenBookQA2018}
Todor Mihaylov, Peter Clark, Tushar Khot, and Ashish Sabharwal. 2018.
\newblock Can a suit of armor conduct electricity? a new dataset for open book
  question answering.
\newblock In \emph{EMNLP}.

\bibitem[{Min et~al.(2019{\natexlab{a}})Min, Chen, Hajishirzi, and
  Zettlemoyer}]{min2019discrete}
Sewon Min, Danqi Chen, Hannaneh Hajishirzi, and Luke Zettlemoyer.
  2019{\natexlab{a}}.
\newblock A discrete hard {EM} approach for weakly supervised question
  answering.
\newblock \emph{arXiv preprint arXiv:1909.04849}.

\bibitem[{Min et~al.(2019{\natexlab{b}})Min, Chen, Zettlemoyer, and
  Hajishirzi}]{min2019knowledge}
Sewon Min, Danqi Chen, Luke Zettlemoyer, and Hannaneh Hajishirzi.
  2019{\natexlab{b}}.
\newblock Knowledge guided text retrieval and reading for open domain question
  answering.
\newblock \emph{arXiv preprint arXiv:1911.03868}.

\bibitem[{Pan et~al.(2019)Pan, Chakravarti, Ferritto, Glass, Gliozzo, Roukos,
  Florian, and Sil}]{pan2019frustratingly}
Lin Pan, Rishav Chakravarti, Anthony Ferritto, Michael Glass, Alfio Gliozzo,
  Salim Roukos, Radu Florian, and Avirup Sil. 2019.
\newblock Frustratingly easy natural question answering.
\newblock \emph{arXiv preprint arXiv:1909.05286}.

\bibitem[{Peters et~al.(2018)Peters, Neumann, Iyyer, Gardner, Clark, Lee, and
  Zettlemoyer}]{peters2018deep}
Matthew~E. Peters, Mark Neumann, Mohit Iyyer, Matt Gardner, Christopher Clark,
  Kenton Lee, and Luke Zettlemoyer. 2018.
\newblock Deep contextualized word representations.
\newblock \emph{arXiv preprint arXiv:1802.05365}.

\bibitem[{Petroni et~al.(2019)Petroni, Rockt{\"a}schel, Lewis, Bakhtin, Wu,
  Miller, and Riedel}]{petroni2019language}
Fabio Petroni, Tim Rockt{\"a}schel, Patrick Lewis, Anton Bakhtin, Yuxiang Wu,
  Alexander~H Miller, and Sebastian Riedel. 2019.
\newblock Language models as knowledge bases?
\newblock \emph{arXiv preprint arXiv:1909.01066}.

\bibitem[{Prager(2006)}]{prager2006open}
John Prager. 2006.
\newblock Open-domain question-answering.
\newblock \emph{Foundations and Trends in Information Retrieval}, 1(2).

\bibitem[{Radford et~al.(2018)Radford, Narasimhan, Salimans, and
  Sutskever}]{radford2018improving}
Alec Radford, Karthik Narasimhan, Tim Salimans, and Ilya Sutskever. 2018.
\newblock Improving language understanding by generative pre-training.

\bibitem[{Radford et~al.(2019)Radford, Wu, Child, Luan, Amodei, and
  Sutskever}]{radford2019language}
Alec Radford, Jeffrey Wu, Rewon Child, David Luan, Dario Amodei, and Ilya
  Sutskever. 2019.
\newblock Language models are unsupervised multitask learners.

\bibitem[{Raffel et~al.(2019)Raffel, Shazeer, Roberts, Lee, Narang, Matena,
  Zhou, Li, and Liu}]{raffel2019exploring}
Colin Raffel, Noam Shazeer, Adam Roberts, Katherine Lee, Sharan Narang, Michael
  Matena, Yanqi Zhou, Wei Li, and Peter~J. Liu. 2019.
\newblock Exploring the limits of transfer learning with a unified text-to-text
  transformer.
\newblock \emph{arXiv preprint arXiv:1910.10683}.

\bibitem[{Rajpurkar et~al.(2016)Rajpurkar, Zhang, Lopyrev, and
  Liang}]{rajpurkar2016squad}
Pranav Rajpurkar, Jian Zhang, Konstantin Lopyrev, and Percy Liang. 2016.
\newblock Squad: 100,000+ questions for machine comprehension of text.
\newblock \emph{arXiv preprint arXiv:1606.05250}.

\bibitem[{Shazeer and Stern(2018)}]{shazeer2018adafactor}
Noam Shazeer and Mitchell Stern. 2018.
\newblock Adafactor: Adaptive learning rates with sublinear memory cost.
\newblock \emph{arXiv preprint arXiv:1804.04235}.

\bibitem[{Talmor et~al.(2019)Talmor, Elazar, Goldberg, and
  Berant}]{talmor2019olmpics}
Alon Talmor, Yanai Elazar, Yoav Goldberg, and Jonathan Berant. 2019.
\newblock olmpics--on what language model pre-training captures.
\newblock \emph{arXiv preprint arXiv:1912.13283}.

\bibitem[{Vaswani et~al.(2017)Vaswani, Shazeer, Parmar, Uszkoreit, Jones,
  Gomez, Kaiser, and Polosukhin}]{vaswani2017attention}
Ashish Vaswani, Noam Shazeer, Niki Parmar, Jakob Uszkoreit, Llion Jones,
  Aidan~N. Gomez, {\L}ukasz Kaiser, and Illia Polosukhin. 2017.
\newblock Attention is all you need.
\newblock In \emph{Advances in Neural Information Processing Systems}.

\bibitem[{Wenzek et~al.(2019)Wenzek, Lachaux, Conneau, Chaudhary, Guzman,
  Joulin, and Grave}]{wenzek2019ccnet}
Guillaume Wenzek, Marie-Anne Lachaux, Alexis Conneau, Vishrav Chaudhary,
  Francisco Guzman, Armand Joulin, and Edouard Grave. 2019.
\newblock Ccnet: Extracting high quality monolingual datasets from web crawl
  data.
\newblock \emph{arXiv preprint arXiv:1911.00359}.

\bibitem[{Yang et~al.(2019)Yang, Dai, Yang, Carbonell, Salakhutdinov, and
  Le}]{yang2019xlnet}
Zhilin Yang, Zihang Dai, Yiming Yang, Jaime Carbonell, Ruslan Salakhutdinov,
  and Quoc~V. Le. 2019.
\newblock {XLNet}: Generalized autoregressive pretraining for language
  understanding.
\newblock \emph{arXiv preprint arXiv:1906.08237}.

\bibitem[{Zhang et~al.(2018)Zhang, Liu, Liu, Gao, Duh, and
  Van~Durme}]{zhang2018record}
Sheng Zhang, Xiaodong Liu, Jingjing Liu, Jianfeng Gao, Kevin Duh, and Benjamin
  Van~Durme. 2018.
\newblock {ReCoRD}: Bridging the gap between human and machine commonsense
  reading comprehension.
\newblock \emph{arXiv preprint arXiv:1810.12885}.

\end{thebibliography}
\bibliographystyle{acl_natbib}

\clearpage

\appendix

\section{Metrics for Natural Questions}
\label{sec:nqeval}

Compared to WebQuestions and TriviaQA, Natural Questions is distributed with a much richer set of annotations: Each question can be annotated either as unanswerable (given the oracle context), with a short answer, or with a yes/no answer; questions in the validation set can be annotated more than once; and some questions have multiple answers (e.g.\ ``Who are the members of the Beatles?'' has four answers).
We consider two variants of Natural Questions.
In both cases, we omit the ``unanswerable'' label and long answers, which are nearly impossible to predict without the oracle context.

The first variant is the standard ``open-domain'' version as used e.g.\ by \cite{lee2019latent,min2019knowledge,min2019discrete,asai2019learning}, where 1) the model is only ever trained to output a single answer; 2) if a question has multiple answers, it is only trained to predict the first answer; 3) any questions with answers longer than five tokens are ignored; 4) answers are normalized before being compared (in the same manner as is typically done for WebQuestions and SQuAD); and 5) a predicted answer is considered correct if it matches any of the answers provided by any of the annotators (e.g.\ ``Ringo Starr'' would be considered a correct answer to ``Who are the members of the Beatles?'').

The second variant closely matches the official evaluation procedure used by the Natural Questions leaderboard, where our model is trained to predict all ground-truth answers and is only considered correct if it predicts \textit{all} answers for any one of the annotators.
As in the official evaluation, we consider questions with fewer than two non-null annotations unanswerable (given the context), but because we cannot predict unanswerability without the context, we only report the recall score.
Further, because our model does not have access to the oracle context, we also normalize predicted and ground-truth answers when comparing them.
The use of multiple possible answers also required minor modification of our text-to-text format.
In this case, we trained the model to output each answer delimited by the text ``answer:'' (for example, ``answer: John Lennon answer: Ringo Starr answer: George Harrison answer: Paul McCartney'').
We then split out each answer from the model's predictions as a postprocessing step before evaluating it against the set of answers provided by each annotation.

\section{Other Things We Tried}
\label{sec:negative}

In the course of undertaking this study, we tried various ideas that ultimately did not improve performance.
We briefly discuss them here.

\paragraph{Continued Pre-Training on Wikipedia} The T5 checkpoints we used were primarily pre-trained on C4, a large and diverse dataset of unstructured web content.
We were interested to see whether we could improve performance by doing further pre-training on data that was better tailored to the tasks we considered.
Since both Natural Questions and TriviaQA source their answers from Wikipedia articles, we experimented with further pre-training on text data from English Wikipedia with the same unsupervised objective (``span corruption'') as was used by T5.
We found that this additional ``in-domain'' pre-training had virtually no effect on performance.
This may be because C4 already contains many articles from Wikipedia and the T5 checkpoints were pre-trained long enough to see plenty of this content.

\paragraph{Pre-Training From Scratch On Wikipedia} Since \textit{all} of the answers to the questions in Natural Questions appeared in Wikipedia, we carried out an additional experiment where we pre-trained T5 from scratch only on data from Wikipedia.
We pre-trained on up to 1 trillion tokens (the same amount the T5 checkpoints were pre-trained on) with the span corruption objective and measured fine-tuned performance after various amounts of pre-training.
Unfortunately, this resulted in dramatically worse performance regardless of the amount of pre-training.
We suspect that this is because Wikipedia is too small and results in detrimental overfitting.

\paragraph{Span-Corruption Pre-Training on Wikipedia Sentences with Salient Spans}

\begin{figure}[h]
    \centering
    \includegraphics[width=0.8\columnwidth]{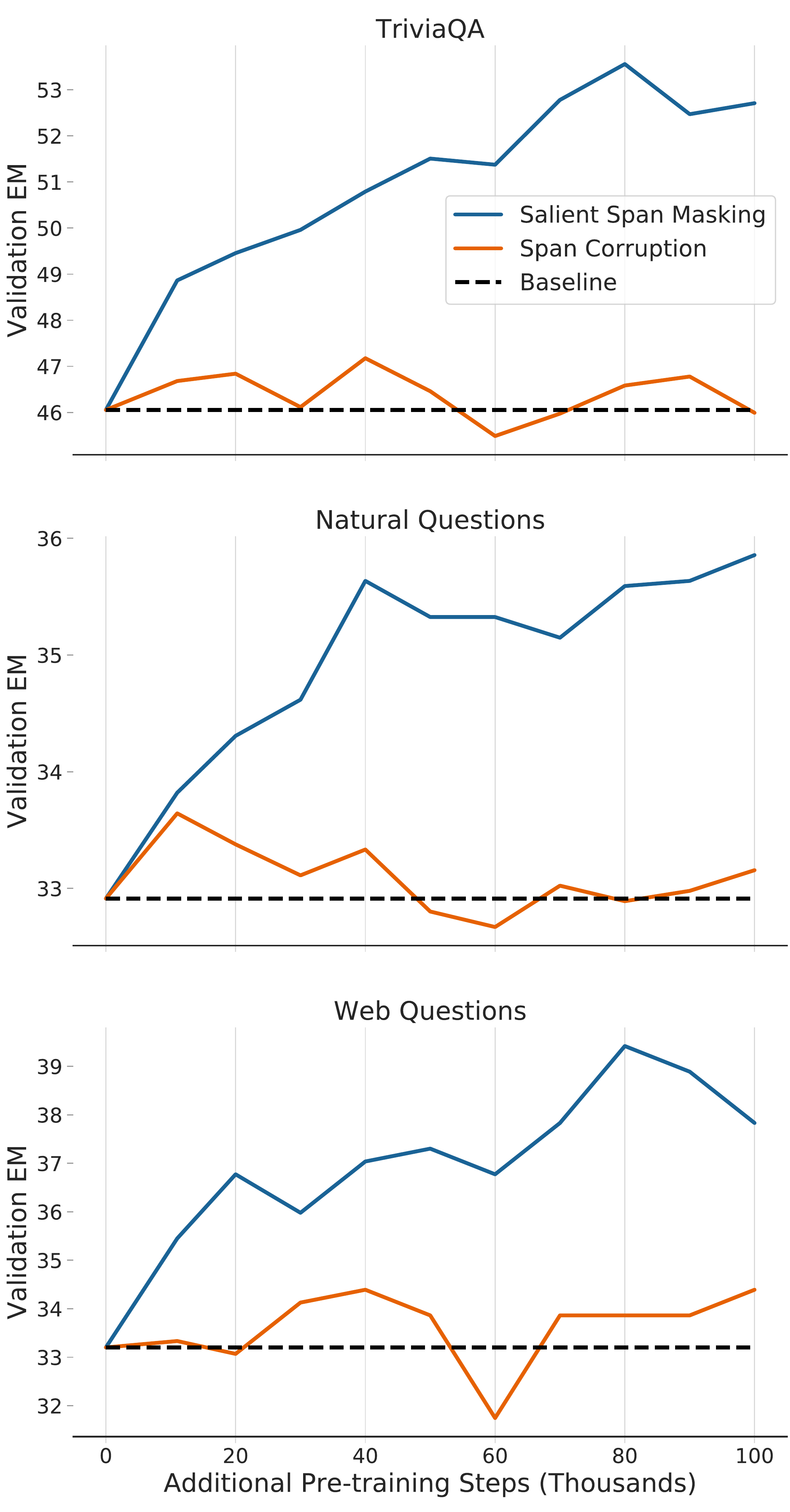}
    \caption{Comparing additional pre-training using either salient span masking (SSM) or span corruption (SC). We further pre-trained T5.1.1-XXL on the Wikipedia sentence dataset from \citet{guu2020realm} with each objective, fine-tuning on a mixture of our three closed-book QA tasks every 10,000 steps. For each fine-tuning run, we report the maximum exact match score achieved on the validation set over 10,000 steps of fine-tuning.}
    \label{fig:ssm_v_sc}
\end{figure}

As described previously, we observed significant performance gains with additional pre-training using ``salient span masking'' (SSM) on the Wikipedia sentence dataset from \citet{guu2020realm} but not when using the standard ``span corruption'' (SC) from \citet{raffel2019exploring} on longer Wikipedia articles.
While SC masks random spans of the input by dropping $15\%$ of its tokens (sampled each epoch) and replacing each consecutive span of dropped tokens with a unique sentinel, SSM specifically masks out one named entity or date in the input sentence.

We were interested in determining whether the gains achieved were attributable to the use of a more task-specific dataset (pre-split into sentences that are known to contain at least one entity) or if the SSM objective itself was critical.
As illustrated in \cref{fig:ssm_v_sc}, the SSM objective is clearly an important ingredient in the improved performance; we saw no significant improvement versus the baseline T5 model when using SC.
 
\paragraph{Fine-Tuning On All Question Answering Tasks} The text-to-text framework used by T5 makes it simple to train multitask models simply by supplying a different task-specific prefix for each task and concatenating all of the constituent datasets.
Since all of the question answering tasks we consider in this study follow the same basic structure, we were hopeful that training on a multitask mixture of Natural Questions, WebQuestions, and TriviaQA would improve performance due to the additional supervised data.
While multitask training improved performance on the Natural Questions by 0.5, it produced slightly worse results on the other tasks.

\paragraph{Randomly Sampling Answers For Natural Questions} In the open-domain variant of Natural Questions, the model is only trained to generate a single answer at a time.
For the results presented in the main text, when a question was annotated with multiple answers, we simply trained the model on the first annotated answer.
We also experimented with sampling a random answer from the set of possible answers for pre-training and found that it did not affect performance.

\end{document}